\pdfoutput=1
\documentclass[11pt,a4paper]{article}
\usepackage{emnlp2021}
\usepackage{times}
\usepackage{latexsym}
\usepackage{amsmath}
\usepackage{booktabs}
\usepackage{tabulary}
\usepackage{graphicx}
\usepackage[multiple]{footmisc}
\usepackage{soul}
\usepackage{fixltx2e}
\usepackage{cleveref}
\usepackage{enumitem}
\usepackage{array}

\usepackage{xcolor}

\crefformat{section}{\S#2#1#3} 
\crefformat{subsection}{\S#2#1#3}
\crefformat{subsubsection}{\S#2#1#3}

\usepackage{array,multirow}
\usepackage{microtype}
\newcommand{\testlay}{Test$_\textsc{lay}$}
\newcommand{\testexpert}{Test$_\textsc{expert}$}
\newcommand{\single}{\textsc{single-sent}}
\newcommand{\double}{\textsc{double-sents}}
\newcommand{\multi}{\textsc{paragraph}}
\newcommand{\indonli}{\textsc{IndoNLI}}
\newcommand{\indolite}{IndoBERT\textsubscript{lite}}
\newcommand{\indolarge}{IndoBERT\textsubscript{large}}

\newcommand{\tabitem}{\textbullet~~}

\usepackage{fdsymbol}

\newcommand{\ui}{$^{1}$}
\newcommand{\kata}{$^{2}$}
\newcommand{\fbk}{$^{3}$}
\newcommand{\oji}{$^{4}$}
\newcommand{\amazon}{$^{5}$}
\title{IndoNLI: A Natural Language Inference Dataset for Indonesian}

\author{Rahmad Mahendra\ui \,
  Alham Fikri Aji\kata \,
  Samuel Louvan\fbk \,\\
  \textbf{Fahrurrozi Rahman\oji\,
  Clara Vania\amazon}\thanks{\ \ Work done while at New York University.}
 \\
  \ui Universitas Indonesia \, \kata Kata.ai Research \, \fbk Fondazione Bruno Kessler\\ \oji University of St Andrews \, \amazon Amazon\
  \\
  \texttt{rahmad.mahendra@cs.ui.ac.id, aji@kata.ai, slouvan@fbk.eu}, \\ \texttt{fr27@st-andrews.ac.uk,  vaniclar@amazon.co.uk}
}

\date{}

\begin{document}
\maketitle
\begin{abstract}

We present IndoNLI, the first human-elicited NLI dataset for Indonesian. 
We adapt the data collection protocol for MNLI and collect $\sim$18K sentence pairs annotated by crowd workers and experts. 
The expert-annotated data is used exclusively as a test set.
It is designed to provide a challenging test-bed for Indonesian NLI by explicitly incorporating various linguistic phenomena such as numerical reasoning, structural changes, idioms, or temporal and spatial reasoning. 
Experiment results show that XLM-R outperforms other pre-trained models in our data.
The best performance on the expert-annotated data is still far below human performance (13.4\% accuracy gap), suggesting that this test set is especially challenging.
Furthermore, our analysis shows that our expert-annotated data is more diverse and contains fewer annotation artifacts than the crowd-annotated data. 
We hope this dataset can help accelerate progress in Indonesian NLP research.

\end{abstract}

\section{Introduction}

\begin{table*}[t]
\small
    \centering
    \begin{tabular}{@{}l@{ ~ }l@{ }c@{ ~ }c@{}}
    \toprule
        \textbf{Premise} & \textbf{Hypothesis} & \textbf{Label} & \textbf{Phenomena} \\

    \midrule
        \textbf{\textsc{Lay-annotated data}} \\
    \midrule    
    
        Seakan tak bisa dipisahkan, dua sahabat itu sama-sama  & Dua sahabat itu selalu bersama-sama. & N & n/a \\
        sedang menggarap proyek musik. & \\
        \textit{(As if they cannot be separated, the two friends} & \textit{(The two friends are always together.)} \\
        \textit{are both working on music projects.)} \\
        
    \midrule
        Meskipun trikomoniasis adalah penyakit yang sangat umum,  & Trikomoniasis bukanlah penyakit & C & n/a \\
        penyakit ini seringkali sulit diketahui. &  yang umum. \\
        \textit{(Although trichomoniasis is very common} & \textit{(Trichomoniasis is not a common }\\
        \textit{disease, it is often difficult to detect.)} & \textit{disease.)} \\
        
    \midrule
        \textbf{\textsc{Expert-annotated data}} \\
    \midrule
        Selanjutnya, \colorbox{lightgray}{dua pemain Arsenal} yang dirasa Ian Wright  & \colorbox{lightgray}{Alexandre Lacazette}\colorbox{pink}{tidak}memiliki  & E & \colorbox{cyan}{MORPH}, \\ 
        \colorbox{pink}{kurang}\colorbox{yellow}{sip}adalah di sektor\colorbox{cyan}{serang.} Mereka adalah  &  performa yang\colorbox{yellow}{baik} sebagai\colorbox{cyan}{penyerang.} & & \colorbox{pink}{NEG}, \\
        \colorbox{lightgray}{Willian dan Alexandre Lacazette}, yang disebutnya buntu. & & & \colorbox{lightgray}{COREF}, \\
        \textit{(Furthermore, \colorbox{lightgray}{two Arsenal players} who Ian Wright felt were} & \textit{(\colorbox{lightgray}{Alexandre Lacazette}\colorbox{pink}{did not}have a} & & \colorbox{yellow}{LEXSEM} \\
        \textit{\colorbox{pink}{not}\colorbox{yellow}{good}enough were in the\colorbox{cyan}{attack}sector. They are} & \textit{\colorbox{yellow}{good}performance as an\colorbox{cyan}{attacker.})}\\
        \textit{\colorbox{lightgray}{Willian and Alexandre Lacazette}, who he calls dead ends.)}\\
         
    \midrule

         Setelah dewasa, Ramlah dinikahi oleh Amr bin  & Ramlah\colorbox{lime}{lebih muda dari pada} & N & \colorbox{lime}{COMP} \\
         Utsman bin Affan. & Amr bin Utsman.\\
         
         \textit{(After growing up, Ramlah was married to } & \textit{(Ramlah was\colorbox{lime}{younger than}} \\
         \textit{Amr bin Uthman bin Affan.}) &  \textit{Amr bin Uthman.)}\\
    
    \bottomrule
    \end{tabular}
    \caption{Examples of premise and hypothesis pairs from \indonli\ (\textbf{E:} Entailment, \textbf{C:} Contradiction, \textbf{N:} Neutral). English translation is provided in the bracket for context. The expert-annotated data is annotated with linguistic phenomena contributing to make the inference. For illustrative purposes, we highlight the sentence chunks that correspond to the specific phenomena (noting that such highlighting is not available in the released dataset).}
    \label{tab:examples}
\end{table*}

Indonesian language or \textit{Bahasa Indonesia} is the 10th most spoken language in the world with more than 190 million speakers.\footnote{\url{https://www.babbel.com/en/magazine/the-10-most-spoken-languages-in-the-world} (Accessed May 2021).} Yet, research in Indonesian NLP is still considered under-resourced due to the limited availability of annotated public datasets.
To help accelerate research progress, \mbox{IndoNLU} \citep{wilie-etal-2020-indonlu} and IndoLEM \citep{koto2020indolem} collect a number of annotated data to benchmark Indonesian NLP tasks. 



In line with their effort, we introduce Indonesian NLI (\indonli),\ a natural language inference dataset for Indonesian. Natural language inference (NLI), also known as \textit{recognizing textual entailment} \citep[RTE;][]{rte} is the task of determining whether a sentence semantically entails another sentence. NLI has been used extensively as a benchmark for NLU, especially with the availability of large-scale English datasets such as the Stanford NLI \citep[SNLI;][]{bowman-etal-2015-large} and the Multi-Genre NLI \citep[MNLI;][]{williams-etal-2018-broad} datasets. Recently, there have been efforts to build NLI datasets in other languages. The most common approach is via translation \citep{conneau-etal-2018-xnli,budur-etal-2020-data}. One exception is the work by \citet{hu-etal-2020-ocnli} which uses a human-elicitation approach similar to MNLI to build an NLI dataset for Chinese (OCNLI). 


Until now, there are two Indonesian NLI datasets available. The first one is WReTE \citep{setya2018semi}, which is created using revision history from Indonesian Wikipedia. The second one, INARTE \citep{INARTE}, is created automatically based on question-answer pairs taken from Web data. Both datasets have relatively small number of examples (400 pairs for WReTE and $\sim$1.5k pairs for INARTE) and only uses two labels (\textit{entailment} and \textit{non-entailment}).
Furthermore, since the hypothesis sentence is generated automatically from the premise, they tend to be so similar that arguably they will not be effective as a benchmark for NLU \citep{Hidayat2021148}. On the other hand, \indonli\ is created using a human-elicited approach similar to MNLI and OCNLI. It consists of $\sim$18K annotated sentence pairs, making it the largest Indonesian NLI dataset to date. 

\indonli\ is annotated by both crowd workers (layperson) and experts. Lay-annotated data is used for both training and testing, while expert-annotated data is used exclusively for testing. Our goal is to introduce a challenging test-bed for Indonesian NLI. Therefore the expert-annotated test data is explicitly designed to target phenomena such as lexical semantics, coreference resolution, idioms expression, and common sense reasoning. 
Table \ref{tab:examples} exemplifies \indonli\ data.

\newpage
We also propose a more efficient label validation protocol. Instead of selecting a consensus gold label from 5 votes as in MNLI data protocol, we incrementally annotate the label starting from 3 annotators. We only add more label annotation if consensus is not yet reached. Our proposed protocol is 34.8\% more efficient than the standard 5 votes annotation.

We benchmark a set of NLI models, including multilingual pretrained models such as XLM-R \citep{conneau-etal-2020-unsupervised} and pretrained models trained on Indonesian text only \citep{wilie-etal-2020-indonlu}. 
We find that the expert-annotated test is more difficult than lay-annotated test data, denoted by lower model performance. The Hypothesis-only model also yields worse results on our expert-annotated test, suggesting fewer annotation artifacts. Furthermore, our expert-annotated test has less hypothesis-premise word overlap, signifying more diverse and creative text. Overall, we argue that our expert-annotated test can be used as a challenging test-bed for Indonesian NLI. 

We publish \indonli\ data and model at \mbox{\url{https://github.com/ir-nlp-csui/indonli}}.


\section{Related Work}
\label{sec:related-work}


\paragraph{NLI Data}
Besides SNLI and MNLI, another large-scale English NLI data which is proposed recently is the Adversarial NLI \citep[ANLI;][]{nie-etal-2020-adversarial}. It is created using a human-and-model-in-the-loop adversarial approach and is commonly used as an extension of SNLI and MNLI.

For NLI datasets in other languages, the Cross-lingual NLI (XNLI) corpus extends MNLI by manually translating sampled MNLI test set into 15 other languages \citep{conneau-etal-2018-xnli}. The Original Chinese Natural Language Inference (OCNLI) is a large-scale NLI dataset for Chinese created using data collection similar to MNLI \citep{hu-etal-2020-ocnli}. Other works contribute to creating NLI datasets for Persian~\citep{farstail} and Hinglish~\citep{khanuja-etal-2020-new}.

Some corpora are created with a mix of machine translation and human participation. The Turkish NLI (NLI-TR) corpus is created by machine-translating SNLI and MNLI sentence pairs into Turkish, which are then validated by Turkish native speakers \citep{budur-etal-2020-data}. For Dutch, \citet{wijnholds-moortgat-2021-sick} introduce the SICK-NL by machine-translating the SICK dataset \citep{marelli2014sick}. It is then manually reviewed to maintain the correctness of the translation.

AmericasNLI, an extension of XNLI to 10 indigenous languages of the Americas, is created with the primary goal to investigate the performance of NLI models in truly low-resource language settings \citep{americasnli}. The dataset is built by translating Spanish XNLI into the target languages, and vice versa, using Transformer-based sequence-to-sequence models.

\paragraph{NLI Analysis}

Research conducted by \citet{tsuchiya-2018-performance, poliak-etal-2018-hypothesis, gururangan-etal-2018-annotation} show that there are hidden biases in the NLI corpus, such as word choices, grammaticality, and sentence length, which allow models to predict the correct label only from the hypothesis.

Several studies also investigate whether NLI models might use heuristics in their learning. Many NLI models still suffer from various  aspects such as antonymy, numerical reasoning, word overlap, negation, length mismatch, and spelling error \citep{naik-etal-2018-stress}, lexical overlap, subsequence and constituent \citep{mccoy-etal-2019-right}, lexical inferences \citep{glockner-etal-2018-breaking} and syntactic structure \citep{poliak-etal-2018-collecting-diverse}. 

Research to analyze which linguistic phenomena are learned by current models has gained interest. This ranges from the definition of the diagnostic test \citep{wang-etal-2018-glue}, the linguistic phenomena \citep{BentivogliCDGLM10}, fine-grained annotation scheme  \citep{analyzingANLI}, to the taxonomic categorization refinement \citep{joshi-etal-2020-taxinli}.

\section{IndoNLI Data Construction}
\label{sec:protocol}

\subsection{Data Source}
Our premise text is originated from three genres: Wikipedia, news, and Web articles. For the news genre, we use premise text from Indonesian PUD and GSD treebanks provided by the Universal Dependencies 2.5 \citep{UD2.5}
and IndoSum \citep{kurniawan2018indosum}, an Indonesian summarization dataset. For the Web data, we use premise text extracted from blogs and institutional websites (e.g., government, university, and school). To maximize vocabulary and topic variability, we set a limit of five text snippets from the same document as premise text. Moreover, the source of premise text covers a broad range of topics including, but not limited to, science, politics, entertainment, and sport. 

In contrast to most previous NLI studies that only use a single sentence as the premise, we use premise text consists of a varying number of sentences, i.e., single-sentence (\single), double-sentence (\double), and multiple sentences (\multi).\footnote{We limit \multi\ to have a maximum 200-word length so that the current pre-trained language model for Indonesian can process it.}

\begin{table}[!t]
\small
\centering
\begin{tabular}{lrrrr}
\toprule
 & \multicolumn{2}{c}{\textbf{Lay}} & \multicolumn{2}{c}{\textbf{Expert}} \\ \midrule

Round & \#pairs & cum-cons & \#pairs & cum-cons \\ \midrule

1st & 4,489 & 74.3\% & 3,008 & 80.0\% \\ 
2nd & 1,155 & 94.7\% & 602 & 93.8\% \\ 
3rd & 237 & 98.0\% & 188 & 99.2\% \\ \midrule

& \multicolumn{4}{c}{\#annotation needed (incl. authoring)} \\ \midrule
\textit{MNLI} & 22,445 & & 15,040 & \\
ours & 14,859 & $\downarrow$ 33.8\% & 9,814 & $\downarrow$ 34.8\% \\ \bottomrule
\end{tabular}
\caption{Number of verified pairs in three-round validation phases. \%cum-cons is the percentage of cumulative pairs with consensus gold label after completing each round. 
We show the efficiency of our proposed three-round validation compared to MNLI-style}
\label{tab:num_annotation_required}
\end{table}

\begin{table*}[!t]
\small
\centering
\begin{tabular}{lrrrrrr}
\toprule
 & \textbf{SNLI}$^{*}$ & \textbf{MNLI}$^{*}$ & \textbf{XNLI$_\textsc{EN}$}$^{*}$ & \textbf{OCNLI}$^{\mathsection}$ & \multicolumn{2}{c}{\textbf{IndoNLI}} \\
 &  &  &  &  & Lay & Expert \\ \midrule

\#pairs in total & 570,152 & 432,702 & 7,500 & 56,525 & 14,728 & 3,008 \\ 
\#pairs validated & 56,941 & 40,000 & 7,500 & 9,913 & 4,489 & 3,008  \\
\% validated per total & 10.0\% & 9.2\% & 100.0\% & 17.5\% & 30.5\% & 100.0\%\\ 
\% pairs with gold label & 98.0\% & 98.2\% & 93.0\% & 98.6\% & 98.0\% & 99.2\%\\ \midrule

individual label = gold label & 89.0\% & 88.7\% & n/a & 87.5\% & 88.8\% & 91.2\%\\ 
individual label = author`s label & 85.8\% & 85.2\% & n/a & 80.8\% & 86.2\% & 89.0\%\\ \midrule

gold label = author`s label & 91.2\% & 92.6\% & n/a & 89.3\% & 90.6\% & 94.0\%\\ 
gold label $\ne$ author`s label & 6.8\% & 5.6\% & n/a & 9.3\% & 7.4\% & 5.2\%\\
no gold label (no 3 labels match) & 2.0\% & 1.8\% & n/a & 1.4\% & 2.0\% & 0.8\%\\ \bottomrule
\end{tabular}
\caption{IndoNLI data labeling agreement, compared to other NLI dataset. $^{*}$The number for SNLI, MNLI, English subset of dev and test split of XNLI (XNLI$_\textsc{EN}$) are copied from original papers~\citep{bowman-etal-2015-large, williams-etal-2018-broad, conneau-etal-2018-xnli}. $^{\mathsection}$For OCNLI, we recalculate the aggregate number from the original paper~\citep{hu-etal-2020-ocnli} that provided detail for 4 different protocols}
\label{tab:agreement}
\end{table*}

\subsection{Annotation Protocol}
To collect NLI data for Indonesian, we follow the data collection protocol used in SNLI, MNLI, and OCNLI. It consists of two phases, i.e., hypothesis writing and label validation. 

The annotation process involves two groups of annotators. We involve 27 Computer Science students as volunteers in the data collection project. All of them are native Indonesian speakers and were taking NLP classes. Henceforth, we refer to them as the \textbf{lay annotators}. 
The other group of annotators, which we call as \textbf{expert annotators} are five co-authors of this paper, who are also Indonesian native speakers and have at least seven years of experience in NLP.

\paragraph{Writing Phase}

In this phase, each annotator is assigned with 100-120 premises. 
For each premise, annotators are asked to write six hypothesis sentences, two for each semantic label (\textit{entailment}, \textit{contradiction}, and \textit{neutral}). This strategy is similar to the MULTI strategy introduced in the OCNLI data collection protocol.\footnote{In OCNLI, three hypothesis sentences per label are created for each premise, resulting in a total of nine hypotheses.} We provide instruction used in the writing phase in Appendix \ref{appendix:annotation-protocol}.

For hypothesis writing involving expert annotators, we further ask the annotators to tag linguistic phenomena required to perform inference on each sentence pair. 
The linguistic phenomena include lexical change, syntactic structure, and semantic reasoning.
We also ask the expert annotators to ensure that the generated premise-hypothesis pairs are reasonably distributed among different linguistic phenomena. Enforcing balanced distribution among all phenomena is challenging because not all phenomena can be applied to the given premise text. We expect this strategy to help us create non-trivial examples covering various linguistic phenomena in the Indonesian language. 

\paragraph{Validation Phase}

We perform label verification for $\sim$30\% and 100\% pairs of lay-authored and expert-authored data, respectively. Our validation process is done through three rounds. In the first round, each pair is relabeled by two other independent annotators. If the label determined by those two annotators is the same as the initial label given by the annotator in the writing phase (author), we assign it as the \textit{gold label}. Otherwise, we move the sentence pair to the second round in which another different annotator provides the label to the data. If any label was chosen by three of the four annotators (i.e., author, two annotators in the first round, and another annotator in the second round), it is assigned as the gold label. If there is no consensus, we proceed to the last round to collect another label from the fourth annotator.

In the MNLI data collection protocol, the goal of the validation phase is to obtain a three-vote consensus from the original label by the author and the labels given by four other annotators. Therefore, the annotation needed under the MNLI protocol is $5N$ for $N$ pairs of data.  For \indonli,\ our three-round annotation process can reduce the number of required annotations to $3N + X + Y$, where $X$ and $Y$ are the numbers of data for the second and third annotation rounds, respectively. 

Table \ref{tab:num_annotation_required} shows that approximately 15K annotations are required to label and validate 3K data in \textsc{Expert} data if we use MNLI-style validation process, while this number can be reduced into less than 10K annotations (34\% more efficient) using our three-round annotation process.\footnote{If we omit the number of annotations in the writing phase and only consider the number of annotations in the validation phase, the efficiency rate is even higher ($>40\%$).} Our proposed strategy requires less annotation cost, which is worthwhile for the NLP research community with a limited budget.


Table \ref{tab:agreement} summarizes our final data, along with a comparison to SNLI, MNLI, XNLI, and OCNLI. Our results are on par with the number reported in SNLI / MNLI and better than OCNLI. About 98\% of the validated pairs have the gold label, suggesting that our dataset is of high quality in general. The annotator agreement for \textsc{Expert} data is higher than \textsc{Lay} data, suggesting that the first is less ambiguous than the latter.

\subsection{The Resulting Corpus}

\begin{table}[t]
    \centering
\small
    \begin{tabular}{@{}l@{}r@{ }r@{ }r@{ }r@{}}
    \toprule
     & Train & Dev & Test$_\textsc{lay}$ & Test$_\textsc{expert}$  \\
     \midrule
    \#entailment & 3476 & 807 & 808 & 1041 \\
    \#contradiction & 3439 & 749 & 764 & 999 \\
    \#neutral & 3415 & 641 & 629  & 944\\
    \midrule
    premise len & 21.0$_{(14.0)}$ & 19.9$_{(10.9)}$ & 20.4$_{(11.6)}$ & 31.1$_{(18.9)}$ \\
    hypothesis len & 7.6$_{(2.9)}$ & 7.7$_{(2.8)}$ & 7.7$_{(3.1)}$ & 9.3$_{(4.2)}$ \\
    \midrule
    \#\single & 8368 & 1784 & 1836 & 1534 \\
    \#\double & 1442 & 336 & 282 & 1043 \\
    \#\multi & 520 & 77 & 83 & 407 \\
    
    \bottomrule
    \end{tabular}
    \caption{IndoNLI corpus statistics. Length is calculated at token-level, with a simple space-delimited tokens. Numbers in the bracket shows the standard deviation.}
    \label{tab:data-stats}
\end{table}

After filtering out premise-hypothesis pairs with no gold labels (no consensus), we ended up with 17,712 annotated sentence pairs.
All expert-annotated pairs possessing gold labels are used as a test set. The lay-annotated pairs are split into development and test sets, such that there is no overlapping premise text between both sets. In the end, we have two separate test sets: \testexpert and \testlay. Sentence pairs that are not included in the validation phase and the lay-annotated pairs without a gold label are used for the training set.\footnote{We use the initial label given by the author.} The number of expert-annotated pairs missing gold labels is extremely small. We excluded them in the distributed corpus. \indonli\ data characteristics is described in the Appendix \ref{appendix:data-statement}~\citep{bender-friedman-2018-data}

The resulting corpus statistic is presented in Table~\ref{tab:data-stats}. We observe that the three semantic labels have a relatively balanced distribution. 
On average, lay-annotated data seems to have a shorter premise and hypothesis length than expert-annotated data. 
In both \textsc{lay} and \textsc{expert} data, single-sentence premises (\single) is the most dominant, followed by \double\ and \multi.

\paragraph{Word Overlap Analysis}

\citet{mccoy-etal-2019-right} show that NLI models might utilize lexical overlap between premise and hypothesis as a signal for the correct NLI label.
To measure this in our data, we use the Jaccard index to measure \textit{unordered} word overlap and the longest common subsequence, LCS, to measure \textit{ordered} word overlap. In addition, we also measure the new token rate (i.e., the percentage of hypothesis tokens not present in the premise) as a proxy to measure token diversity in the hypothesis. Table \ref{tab:analysis_by_similarity} shows our results. Regarding the Jaccard index, \testexpert\ has an overall lower similarity than \testlay\, and the two have higher similarity for pairs with entailment labels than the other labels. \testexpert\ also has a lower LCS similarity score than \testlay, suggesting that the expert annotators use different wording or sentence structure in the hypothesis. In terms of the new token rate, we find that \testexpert has a higher rate than \testlay.\ This indicates that, in general, expert annotators use more diverse tokens than lay annotators when generating hypotheses.

\begin{table}[t]
    \small
    \centering
    \begin{tabular}{lrrr}
    \toprule
      & Jaccard & LCS & New token rate \\
    \midrule
    \testlay\\
    \midrule
    Entailment & 31.8 &	71.4 & 16.7 \\
    Contradiction & 28.6 & 66.7 & 25.0 \\
    Neutral &	21.1 &	54.5 & 37.5 \\
    \midrule
    \testexpert\\
    \midrule
    Entailment & 21.1 & 60.0 & 30.0 \\
    Contradiction & 20.8 & 62.5 & 28.6 \\
    Neutral & 15.1 & 44.4 & 46.2 \\ 
    \bottomrule
    \end{tabular}
    \caption{Word overlap between premise and hypothesis in \testlay\ vs. \testexpert.\ }
    \label{tab:analysis_by_similarity}
\end{table}

\section{Experiments}

We experiment with several neural network-based models to evaluate the difficulty of our corpus. As our baseline, we use a continuous bag of words (CBoW) model, initialized with Indonesian fastText embeddings \citep{bojanowski-etal-2017-enriching}. The others are Transformers-based models: multilingual BERT \citep{devlin-etal-2019-bert}, XLM-R \citep{conneau-etal-2020-unsupervised}, and two pre-trained models for Indonesian, IndoBERT and IndoBERT-lite \citep{wilie-etal-2020-indonlu}. The first two are multilingual models which are trained on multilingual data, which includes Indonesian text. IndoBERT uses BERT large architecture with 335.2M parameters, while IndoBERT-lite uses ALBERT \citep{Lan2020ALBERT} architecture with fewer number of parameters (11.7M).\footnote{We use indobert-large-p2 and indobert-lite-base-p2 models for our experiments.} 

\paragraph{Training and Optimization}

We use Adam \cite{DBLP:journals/corr/KingmaB14} optimizer for training all models. For the experiment with the CBoW model, we use a learning rate $1 \times 10^{-4}$, batch size of $8$, and dropout rate $0.2$. For experiments with Transformers models, we perform hyperparameter sweep on the learning rate $\in \{1 \times 10^{-5}, 3 \times 10^{-5}, 1 \times 10^{-6}\}$ and use batch size of $16$. Each model is trained for 10 epochs with early stopping, with three random restarts. 

For all of our experiments, we use the \texttt{jiant} toolkit \cite{phang2020jiant}, which is based on Pytorch \citep{pytorch2019} and HuggingFace Transformers \citep{wolf-etal-2020-transformers}. We use NVIDIA V100 Tensor Core GPUs for our experiments. More details regarding training time can be found in the Appendix \ref{appendix:training-time}.

\section{Results and Analysis}

\begin{table}[t]
    \centering
    \small
    \begin{tabular}{lrrr}
    \toprule
     & Dev & Test$_\textsc{lay}$ & Test$_\textsc{expert}$ \\
    \midrule
    Human &  - & \textit{85.1$_{(0.0)}$} & \textit{89.1$_{(0.0)}$}\\
    \midrule
    Majority &  35.7$_{(0.0)}$ & 36.7$_{(0.0)}$ & 34.9$_{(0.0)}$\\
    CBoW &  54.7$_{(0.2)}$ & 50.1$_{(0.5)}$ & 36.9$_{(0.3)}$\\
    \midrule
    IndoBert$_\text{lite}$ & 76.2$_{(0.5)}$ & 74.1$_{(0.2)}$ & 58.9$_{(0.9)}$ \\ 
    IndoBert$_\text{large}$ & 78.7$_{(0.4)}$ & 77.1$_{(0.5)}$ & 61.5$_{(2.2)}$ \\ 
    mBERT & 76.2$_{(0.8)}$& 72.5$_{(0.4)}$ & 57.3$_{(0.8)}$ \\ 
    XLM-R & \textbf{85.7$_{(0.4)}$} & \textbf{82.3$_{(0.3)}$} & \textbf{70.3$_{(1.0)}$} \\
    \bottomrule
    \end{tabular}
    \caption{Average model accuracy on the development and test sets over three random restarts. Numbers in the bracket shows the standard deviation. See Appendix \ref{appendix:human-baseline} for the detail about human baseline.}
    \label{tab:main-result}
\end{table}

\subsection{Model Performance}

Table \ref{tab:main-result} reports the performance of all models on the \indonli\ dataset, along with human performance on both test sets \citep{nangia-bowman-2019-human}. CBoW model gives moderate improvements over the majority baseline. However, as expected, its performance is still below the Transformers model performance. We observe that \indolite\ obtains comparable or better performance than mBERT. \indolarge\ has better performance than \indolite,\ but worse than XLM-R. This is interesting since one might expect that the Transformers model trained only on Indonesian text will give better performance than a multilingual Transformers model. One possible explanation is because the size of Indonesian pretraining data in XLM-R is much larger than the one used in \indolarge\ (180GB vs. 23GB uncompressed). This finding is also in line with IndoNLU benchmark results \citep{wilie-etal-2020-indonlu}, where XLM-R outperforms \indolarge\ on several tasks.

In terms of difficulty, it is evident that \testexpert\ is more challenging than \testlay\ as there is a large margin of performance (up to 16\%) between \testexpert\ and \testlay\ across all models. 
We also see larger human-model gap in \testexpert\ (18.8\%) compared to \testlay (2.8\%). This suggests that \indonli\ is relatively challenging for all the models, as there is still room for improvements for \testlay\ and even more for \testexpert.\ 

\begin{table}[t]
    \small
    \centering
    \begin{tabular}{lrr}
    \toprule
      & \testlay & \testexpert \\
    \midrule
    \textit{By Label}\\
    \midrule
    Entailment &	81.4 &	56.6 \\
    Contradiction & 82.7 & 63.3 \\
    Neutral &	83.1 &	90.1 \\
    \midrule
    \textit{By Premise Sentence Type}\\
    \midrule
    \single & 82.3 & 70.9 \\
    \double & 83.3 &	69.0 \\
    \multi & 79.5 &	65.3 \\ 
    \bottomrule
    \end{tabular}
    \caption{Performance comparison of lay and expert data based on the NLI label and  the premise type.}
    \label{tab:analysis_by_label_premise_type}
\end{table}

\paragraph{Analysis by Labels and Premise Type} 
We compare the performance based on the NLI labels and premise type between \testlay\ and \testexpert\ (Table \ref{tab:analysis_by_label_premise_type}). We observe that the accuracy across labels is similar on the lay data, while on \testexpert,\ the performance between labels is substantially different.  Overall, the neutral label is relatively easier to predict than other labels for both \testlay\ and \testexpert.\ Contradiction and entailment labels for \testexpert\ are considerably more difficult than \testlay.\ For the performance based on the premise sentence type, there is no substantial accuracy difference between \single\ and \double\ for both \testexpert\ and \testlay.\ When moving to multiple-sentence premise type (\multi), we observe a large drop in performance for both \testexpert\ and \testlay.\ 

\subsection{Annotation Artifacts}

\begin{table}[t]
    \small
    \centering
    \begin{tabular}{lrrr}
    \toprule
     & Dev & \testlay & \testexpert \\
    \midrule
    \indolite\  & 57.6$_{(0.4)}$ & 56.7$_{(0.5)}$ & 45.9$_{(0.4)}$ \\
    \indolarge\ & 60.3$_{(0.9)}$ & 59.5$_{(1.2)}$ & 45.7$_{(1.7)}$ \\
    mBERT  & 57.5$_{(0.8)}$& 56.9$_{(1.1)}$ & 44.5$_{(1.0)}$ \\
    XLM-R & \textbf{60.5$_{(0.5)}$} & \textbf{59.8$_{(1.0)}$} & \textbf{46.0$_{(0.8)}$} \\
    \bottomrule
    \end{tabular}
    \caption{Hypothesis-only baseline results.}
    \label{tab:hyp-only}
\end{table}

\paragraph{Hypothesis-only Models}

\begin{table}[t]
    \small
    \centering
    \begin{tabular}{llcrr}
    \toprule
    \multicolumn{2}{c}{Word} & Label & PMI & Count \\
    \midrule
    \textsc{\textbf{Lay}}\\
    \midrule
    salah & \textit{wrong} & E & 0.72 & 96/160 \\
    sekitar & \textit{around} & E & 0.72 & 40/60 \\
    suatu & \textit{something} & E & 0.58 & 32/53 \\
    \midrule
    bukan & \textit{no}  & C & 1.28 & 279/324 \\
    tidak & \textit{no} & C &1.24 & 2319/2905 \\
    apapun & \textit{anything} & C &1.20 & 67/70 \\
    \midrule
    selain & \textit{aside from} & N & 1.08 & 66/80 \\
    juga & \textit{also} & N & 1.05 & 109/146 \\
    banyak & \textit{a lot} & N & 0.94 & 291/452 \\
    \midrule
    \textsc{\textbf{Expert}}\\
    \midrule
    beberapa & \textit{some} & E & 0.65 & 40/65 \\
    dapat & \textit{can} & E & 0.50 & 44/84 \\
    ajaran & \textit{doctrine} & E & 0.48 & 12/17 \\
    \midrule
    tidak & \textit{no}  & C & 0.84 & 205/329 \\
    kurang & \textit{less} & C & 0.50 & 23/40 \\
    didirikan & \textit{established} & C & 0.49 & 14/21 \\
    \midrule
    banyak & \textit{a lot} & N & 0.69 & 54/90 \\
    ia & \textit{he/she} & N & 0.67 & 32/50 \\
    juga & \textit{also} & N & 0.63 & 37/62 \\
    \bottomrule
    \end{tabular}
    \caption{Top 3 PMI values between words and label on lay and expert data. \textbf{E}: Entailment, \textbf{C}: Contradiction, \textbf{N}: Neutral.}
    \label{tab:pmi}
\end{table}

\citet{poliak-etal-2018-hypothesis} propose a hypothesis-only model as a baseline when training an NLI model to investigate statistical patterns in the hypothesis that may reveal the actual label to the model. Table \ref{tab:hyp-only} shows results when we only use hypothesis as input to our models. On \testlay\ split, our best performing model achieves $\sim$60\%, slightly lower than other NLI datasets (MNLI $\sim$ 62\%; SNLI $\sim$ 69\%, and OCNLI $\sim$66\%). We see much lower performance on the \testexpert\ split, with performance reduction up to 14\%. This result indicates that our protocol for collecting \testexpert\ effectively reduces annotation artifacts that \testlay\ has. However, since we do not use expert-written examples for training, \testexpert\ may have different artifacts that our models do not learn.

\paragraph{PMI Analysis}
We compute Pointwise Mutual Information (PMI) to see the discriminative words for each NLI label (Table \ref{tab:pmi}). Manual analysis suggests that some words are actually part of multi-word expression~\citep{suhardijanto-etal-2020-framework}. For example, the word \textit{salah} is actually part of the expression \textit{salah satu} which means \textit{one of}. In general, we observe that the PMI values in lay data are relatively higher than expert data, indicating that the expert data has better quality and is more challenging. For contradiction label, it is dominated by negation words (e.g., \textit{bukan}, \textit{tidak} ). However, for expert data, only one negation word presents in the top 3 words, while for lay data, all top three words are negation words. This suggests that our annotation protocol in constructing expert data is effective in reducing these particular annotation artifacts.


\begin{table}[t]
    \small
    \centering
    \begin{tabular}{lrrr}
    \toprule
    & Dev & \testlay & \testexpert \\
    \midrule
    \textit{Zero-shot} & 80.8$_{(0.0)}$ & 78.3$_{(0.0)}$ & 73.8$_{(0.0)}$ \\
    \textit{Translate-train} & 82.8$_{(1.1)}$ & 80.4$_{(0.9)}$ & \textbf{75.7$_{(0.8)}$} \\
    \textit{Translate-train-s} & 79.3$_{(0.9)}$ & 76.7$_{(0.5)}$ & 71.1$_{(0.2)}$ \\
    \midrule
    \indonli\ & \textbf{85.7$_{(0.4)}$} & \textbf{82.3$_{(0.3)}$} & 70.3$_{(1.0)}$ \\
    \bottomrule
    \end{tabular}
    \caption{Comparison with \textit{zero-shot} and \textit{translate-train} approaches.}
    \label{tab:cross-lingual}
\end{table}

\section{Cross-Lingual Transfer Performance}

Prior work \citep{conneau-etal-2018-xnli,budur-etal-2020-data} has demonstrated the effectiveness of cross-lingual transfer when training data in the target language is not available. To evaluate the difficulty of our test sets in this setting, we experiment with two cross-lingual transfer approaches, zero-shot learning, and translate-train. In this experiment, we only use XLM-R as it obtains the best performance in our NLI evaluation.

In the \textit{zero-shot learning}, we employ an XLM-R model trained using a concatenation of MNLI training set and XNLI validation set, which covers 15 languages in total.\footnote{We use a pre-trained model distributed by HuggingFace: \url{https://huggingface.co/joeddav/xlm-roberta-large-xnli}.} In the \textit{translate-train} setting, we machine-translate MNLI training and validation sets into Indonesian and fine-tune the pre-trained XLM-R on the translated data. Our English to Indonesian machine translation system uses the standard Transformer architecture~\cite{vaswani2017attention} with 6 layers of encoders and decoders. Following~\citet{guntara-etal-2020-benchmarking}, we train our translation model on a total of 13M pairs of multi-domain corpus from news articles, religious texts, speech transcripts, Wikipedia corpus, and back-translation data from OSCAR~\cite{ortiz-suarez-etal-2020-monolingual}. We use Marian~\cite{mariannmt} toolkit to train our translation model.

\begin{table*}[t]
    \small
    \centering
    \begin{tabular}{llrrrrccc}
    \toprule
    \multicolumn{2}{c}{Inference tags} & \multicolumn{4}{c}{\#pairs (E:C:N)} & \multicolumn{3}{c}{Accuracy (\%)} \\
    
    \multicolumn{6}{c}{} & 
    \indolarge & XLM-R & \textit{translate-train} \\
    \midrule
    
Morphological derivation & \textsc{MORPH} & 96 & (47 & 31 & 18) & 69.4$_{(3.2)}$ &	79.2$_{(1.8)}$ & 84.4$_{(1.0)}$ \\
Syntactic structure reordering & \textsc{STRUCT} & 100 & (53 &	36 & 11) &	60.7$_{(2.5)}$ & 73.3$_{(1.5)}$ &	79.0$_{(2.6)}$ \\
Lexical subsequence & \textsc{LSUB} &	99 & (48 &	42 & 9) &	66.7$_{(7.1)}$	& 74.4$_{(0.6)}$	& 85.2$_{(3.1)}$ \\
Negation & \textsc{NEG}	& 75 & (11 & 55 & 9) &	71.1$_{(0.8)}$ &	71.6$_{(0.8)}$	& 79.1$_{(3.4)}$ \\
Coordinating conjunction & \textsc{COORD} & 38 & (14 & 13 &	11) &	69.3$_{(8.5)}$ &	80.7$_{(1.5)}$ &	85.1$_{(4.0)}$ \\
Logical quantification & \textsc{QUANT}	& 59 & (18 & 20 & 21) &	68.9$_{(2.6)}$	& 75.7$_{(1.0)}$	& 75.1$_{(3.5)}$ \\
Numerical \& math reasoning & \textsc{NUM} &	120 & (40 &	43	& 37) &	45.0$_{(2.2)}$ &	58.3$_{(2.2)}$ &	65.0$_{(0.0)}$ \\
Comparative \& superlative & \textsc{COMP} & 51 & (12 &	15 &	24) &	59.5$_{(4.1)}$	& 64.1$_{(3.0)}$	& 59.5$_{(7.9)}$ \\
Lexical semantics & \textsc{LEXSEM}	& 166 & (68 &	73 &	25) &	58.4$_{(4.2)}$ &	71.3$_{(0.9)}$ &	76.9$_{(3.0)}$ \\
Idiomatic expression & \textsc{IDIOM} & 28 &	(12 &	3 &	13) &	59.5$_{(5.5)}$	& 69.0$_{(5.5)}$ &	78.6$_{(3.6)}$ \\
Anaphora \& coreference & \textsc{COREF} & 70 &	(29 &	23 &	18) &	64.8$_{(5.0)}$ &	74.3$_{(1.4)}$ &	74.8$_{(2.2)}$ \\
Spatial reasoning & \textsc{SPAT} &	37 & (11 &	8 &	18) &	45.9$_{(2.7)}$ &	57.7$_{(1.6)}$ &	68.5$_{(5.6)}$ \\
Temporal expression \& reasoning & \textsc{TEMP} & 68 &	(15 &	20 &	33) &	56.4$_{(8.9)}$ &	68.1$_{(2.2)}$ &	69.6$_{(3.7)}$ \\
Common-sense reasoning & \textsc{CS} & 105 & (42 &	18 &	45) &	55.6$_{(3.1)}$	& 63.8$_{(1.6)}$	& 70.2$_{(1.5)}$ \\
World knowledge & \textsc{WORLD} & 70 & (16 &	20 &	34) & 67.1$_{(3.8)}$	& 73.3$_{(0.8)}$	& 69.0$_{(1.6)}$ \\

    \bottomrule
    \end{tabular}
    \caption{Inference tags examined in \indonli\ diagnostic set and model performance measured on pairs annotated with tag}
    \label{tab:diagnostic-test}
\end{table*}

Translate-train outperforms a model trained on our training data (\indonli)\ on \testexpert.\ Translate-train obtains the best performance with 5.4 points over the \indonli\ model. We further investigate if the performance gap comes from the larger training data used for training our translate-train model. We train another model (\textit{translate-train-s}), using a subset of translated training data such that the size is comparable to \indonli\ training data. We find that the model performance using the same training data is also higher, although the gap is smaller (0.8 points). Since we do not include expert data in the \indonli\ training data, this result indicates that the translated training data might contain more examples with similar characteristics with examples in \testexpert\ than \indonli\ training data. Overall, we observe that the best performance on our \testexpert\ is still relatively low (e.g., 75.7 compared to the best performance in OCNLI, 78.2), indicating that the test set is still challenging.

\section{Linguistic Phenomena in \testexpert\ }

To investigate natural language challenges in \indonli\ data, 
we perform an in-depth analysis of linguistic phenomena and task accuracy breakdown~\citep{linzen-2020-accelerate} on our test-bed. Specifically, we examine the distribution of inference categories in \testexpert\ data and investigate which category the models succeed or fail.

We curate a subset of 650 examples from \testexpert\ as the diagnostic dataset.\footnote{We use the same examples to evaluate human baseline} To annotate the diagnostic set with linguistics phenomena, we ask one expert annotator (who is not the example's author) to review the inference categories tagged by the expert-author when creating premise-hypotheses pairs~\citep{analyzingANLI}. The annotation is multi-label, in which a premise-hypotheses pair can correspond to more than one natural language phenomenon.

Our annotation scheme incorporates 15 types of inference categorization. They include a variety of linguistic and logical phenomena and may require knowledge beyond text. The definition for inference tags examined in diagnostic set is stated in the Table \ref{tab:phenomena} in the Appendix \ref{appendix:phenomena}. We provide the tag distribution in diagnostic set and also report the performance of \indolarge,\ XLM-R, and \textit{translate train} models on the curated examples, as shown in Table \ref{tab:diagnostic-test}.

We see that many premise-hypothesis pairs in \indonli\ diagnostic set apply lexical semantics (e.g., synonyms, antonyms, and hypernyms-hyponyms) or require common-sense knowledge (i.e., a basic understanding of physical and social dynamics) to make an inference. Pairs with \textsc{NUM} tag also occur with high frequency in our data, whereas the challenges of idiomatic expression is less prevalent. Few phenomena are evenly distributed among labels, e.g., \textsc{NUM}, \textsc{QUANT}, and \textsc{COORD}. On the other hand, there is only a small proportion of pairs in \textsc{LSUB} and \textsc{STRUCT} categories which have \textit{neutral} labels. Many examples of \textsc{NEG} unsurprisingly have \textit{contradiction} label.


Our analysis on diagnostic set shows that the models can handle examples tagged with morpho-syntactic categories or boolean logic well. \textsc{COORD} and \textsc{MORPH} are among 2 tags with highest performance for XLM-R. The translate-train model achieves 85\% on \textsc{LSUB}, indicating that our data is considerably robust with respect to syntactic heuristics, such as lexical overlap and subsequence \citep{mccoy-etal-2019-right}. On the other hand, all models also have decent accuracy on inference pairs with negation; the performance remains stable in three different models (more than 70\%).

In contrast, the hardest overall categories appear to be \textsc{NUM} and \textsc{COMP}, indicating that models struggle with arithmetic computation, reasoning about quantities, and dealing with comparisons~\citep{ravichander-etal-2019-equate,roy-etal-2015-reasoning}. In addition, models find it difficult to reason about temporal and spatial properties, as the accuracy for examples annotated with \textsc{TEMP} and \textsc{SPAT} are also inadequate. Commonsense reasoning is shown as another challenge for NLI model, suggesting that there is still much room for improvement for models to learn tacit knowledge from the text.





\section{Conclusion}

We present \indonli,\ the first human-elicited NLI data for Indonesian. The dataset is authored and annotated by crowd (\textit{\textit{lay data}}) and expert annotators (\textit{\textit{expert data}}). \indonli\ includes nearly 18K sentence pairs, makes it the largest Indonesian NLI dataset to date. We evaluate state-of-the-art NLI models on \indonli,\ and find that our dataset, especially the one created by \textit{expert} is challenging as there is still a substantial human-model gap on \testexpert.\
The expert data contains more \textit{diverse hypotheses} and \textit{less annotation artifacts} makes it ideal for testing models beyond its normal capacity (\textit{stress test}). Furthermore, our qualitative analysis shows that the best model struggles in handling linguistic phenomena, particularly in numerical reasoning and comparatives and superlatives. We expect this dataset can contribute to facilitating further progress in Indonesian NLP research. 

\section{Ethical Considerations}
\label{sec:ethics}

\indonli\ is created using premise sentences taken from Wikipedia, news, and web domains. These data sources may contain harmful stereotypes, and thus models trained on this dataset have the potential to reinforce those stereotypes. We argue that additional measurement on the potential harms introduced by these stereotypes is needed before using this dataset to train and deploy models for real-world applications.

\section*{Acknowledgments}

We would like to thank Kerenza Doxolodeo, Theresia Veronika Rampisela, and Ajmal Kurnia for their contribution in preparing unannotated data set and assisting the annotation process. We also thank the student annotators, without whom this work would not have been possible.

RM's work on this project was financially supported by a grant from Program Kompetisi Kampus Merdeka (PKKM) 2021, Faculty of Computer Science, Universitas Indonesia.

CV's work on this project at New York University was financially supported by Eric and Wendy Schmidt (made by recommendation of the Schmidt Futures program) and Samsung Research (under the project \textit{Improving Deep Learning using Latent Structure}) and benefitted from in-kind support by the NYU High-Performance Computing Center. This material is based upon work supported by the National Science Foundation under Grant No. 1922658. Any opinions, findings, and conclusions or recommendations expressed in this material are those of the author(s) and do not necessarily reflect the views of the National Science Foundation.


\bibliographystyle{acl_natbib}
\bibliography{anthology,emnlp2021}

\appendix

\clearpage
\section{Data Statement}
\label{appendix:data-statement}

\subsection{Curation Rationale}

\indonli\ is the first human-elicited natural language inference (NLI) dataset for Indonesian. It is created to foster research in Indonesian NLP, especially for NLI.

The dataset consists of 17,712 premise-hypothesis pairs and is divided into training, development, and test splits. We curate two separate test sets, \testlay,\ which is authored and annotated by 27 students, and \testexpert,\ which is authored and annotated by experts. The students were offered the compensation whose the rate is similar to the wage for a research assistant in Faculty of Computer Science, Universitas Indonesia. All authors and annotators are Indonesian native speakers.
\subsection{Language Variety}

The premise sentences used to create \indonli\ are taken from three sources: Wikipedia, news, and web domain. For the news text, we use premise sentences from the Indonesian PUD\footnote{\url{https://github.com/UniversalDependencies/UD\_Indonesian-PUD}} and GSD\footnote{\url{https://github.com/UniversalDependencies/UD\_Indonesian-GSD}} treebanks provided by the Universal Dependencies 2.5 \citep{UD2.5} and IndoSum dataset \citep{kurniawan2018indosum}. For the web domain, we collect premise sentences from blogs and institutional websites (e.g., government, university, and school). Our manual analysis shows that most of the sentences are written in standard written Indonesian.

\subsection{Speaker Demographic}

All \indonli\ authors are Indonesian native speakers. Lay authors are undergraduate students who have taken NLP class, while expert authors consist of 2 Ph.D. students and 3 researchers with at least 7 years experience in NLP. Besides this information, we do not collect any other demographic information of the authors.

\subsection{Annotator Demographic}

The authors of \indonli\ are also the annotators.

\subsection{Speech Situation}

Each hypothesis in \indonli\ is written based on premise sentence(s) taken from Wikipedia, news, or web articles.

\subsection{Text Characteristics}

The premise text in \indonli\ can be categorized into three groups: single-sentence (\single),\ double-sentence (\double),)\ and multiple-sentence (\multi).\

\subsection{Recording Quality}
N/A

\subsection{Other}
N/A

\subsection{Provenance Appendix}
N/A

\vspace{10mm}

\section{Training Time}
\label{appendix:training-time}
\begin{table}[h]
    \centering
    \small
    \begin{tabular}{l r r}
    \toprule
         \multirow{2}{*}{\textbf{Model}}&  \multicolumn{2}{c}{\textbf{Training Data}}\\
         & IndoNLI & Indo\_XNLI  \\
         & (hrs) & (days) \\
         \midrule
        mBERT & $\pm 3$  & $\pm1$ \\
        XLM-R & $\pm10$  & $\pm4$ \\
        \indolarge\ & $\pm3$  & $\pm1$  \\
        \indolite & $\pm6$   & $\pm3$ \\
    \bottomrule        
    \end{tabular}
    \caption{Training time for a single run on each model on a particular training data.}
    \label{tab:my_label}
\end{table}

\vspace{10mm}

\section{Determining Human Baseline}
\label{appendix:human-baseline}

We follow the procedure in \citet{nangia-bowman-2019-human} to measure human baselines performance on
\indonli.\ We hire 3 Indonesian native speakers who do not participate in the data collection process. We provide them with 15 examples of labeled premise-hypothesis pairs (5 pairs for each label). We also tailor a short prompt explaining the NLI task definition. After reviewing the examples and prompt, the annotators are then given a stratified sample of 450 and 650 examples from \testlay and \testexpert,\ respectively. The gold label for a total of 1,100 examples are concealed, and the annotators are asked to perform labeling experiment. We compute the majority label from them and compare that against the gold label in \indonli\ test data to obtain accuracy. For pairs with no majority label, we use the most
frequent label from \indonli\ test data (\textit{entailment}).

\clearpage

\onecolumn

\section{\indonli\ Writing Instruction}
\label{appendix:annotation-protocol}

\begin{small}
\begin{center}
\begin{tabular}{p{0.85\textwidth}}
\\
\bottomrule\bottomrule
\\
In this task, given a premise text consisting of one or more sentences, you are asked to write six different hypothesis sentences, two for each label (\textit{entailment, contradiction, and neutral}). \\ 
\\

A premise-hypothesis pair is annotated with \textbf{\textit{entailment}} label if it can be concluded that the hypothetical text is correct based on the information contained in the premise text. It is annotated with \textbf{\textit{contradiction}} label is if it can be concluded that the hypothesis text is wrong based on the information contained in the premise text. Otherwise, the label is \textbf{\textit{neutral}}; in other words, based on the information contained in the premise text, the truth of the hypothesis text cannot be determined (not enough information).\\

\\
Please make sure that each hypothesis sentence satisfies the following criteria: \\
\tabitem It consists of one sentence and not multiple sentences, \\
\tabitem It contains some keywords present in the premise text, \\
\tabitem It is grammatical according to Indonesian grammar. \\
\\
Some strategies that you can apply when writing the hypothesis sentence including, but not limited to:
\begin{enumerate}
\item \textit{Word deletion} \newline
Delete one or more words from the premise text.

\item \textit{Word addition} \newline
Add one or more words to the premise text. For example, you can add adjectives, negation words, etc.

\item \textit{Lexical change} \newline
Replace one or more words from premise text with their synonym, antonym, hypernym, or hyponym.

\item \textit{Paraphrase} \newline
Write premise text with your own words.

\item \textit{Structural change} \newline
Change the structure of the premise text. For example, you can change the active voice into passive voice or change the order of the sub-sentence in the premise text.

\item \textit{Reasoning} \newline
Apply reasoning to the given premise text to write a hypothesis sentence, such that the reasoning skill is needed when deciding the correct entailment label. For example, you can use numerical reasoning or commonsense knowledge.
\end{enumerate}

If you are given a premise text which consists of multiple sentences, you \textit{should not} write a hypothesis sentence that is \textit{identical} to one of the sentences in the premise text. \\

\\
\bottomrule\bottomrule
\\
\end{tabular}

\noindent\normalsize{Figure 1: This is the instruction given to lay authors for writing hypothesis sentences.}\\\bigskip

\end{center}
\end{small}

\vspace{5mm}

\section{\indonli\ Validation Instruction}

\begin{small}
\begin{center}
\begin{tabular}{p{0.85\textwidth}}
\\
\bottomrule\bottomrule
\\
In this task, you will be given a set of sentence pairs. For each sentence pair: 
\begin{enumerate}
    \item Check if they are free of errors such as ungrammatical, incomplete, or have wrong punctuation.
    \item Fix any error that is found in one or both sentences.
    \item Pick the correct semantic label for the sentence pair.
\end{enumerate}

\\
\bottomrule\bottomrule
\\
\end{tabular}

\noindent\normalsize{Figure 2: This is the instruction given to annotators for labeling each sentence pair.}\\\bigskip

\end{center}
\end{small}

\vspace{5mm}

\newpage

\section{The Linguistic Phenomena Examined in IndoNLI}
\label{appendix:phenomena}

\begin{table*}[ht!]
\small
    \centering
    \begin{tabular}{@{}l@{ ~ }l@{ }l}
    \toprule
        \textbf{Our Tag} & \textbf{Description} & \textbf{Similar Tag in Related Work}\\

    \midrule
        MORPH &	Transforming the word form by applying morphological & Lexical:Verbalization \citep{BentivogliCDGLM10}\\
        & derivation. For example, a noun into verb (verbalization) or & \\
        & vice versa (nominalization). & \\
    
        \midrule
        STRUCT & Reordering the structure of arguments in the premise, & Alternations \citep{wang-etal-2018-glue}\\
        & e.g. changing active into passive voices. & Syntactic \citep{joshi-etal-2020-taxinli}\\
    
    
         \midrule
        LSUB & Syntactic heuristics, i.e., word overlap and lexical subsequence. & \citep{mccoy-etal-2019-right}\\
        & This phenomena is captured in the data whose the hypothesis is & Word overlap \citep{naik-etal-2018-stress}\\
        & obtained by deleting or adding the words without changing & \\
        & the sentence structure & \\
    

    \midrule
        NEG	& The use of negation words, e.g., \textit{tidak}, \textit{bukan} & \citep{hossain-etal-2020-analysis} \\
        && Negation \citep{kim-etal-2019-probing}\\
        && Negation \citep{naik-etal-2018-stress}\\
        && Negation \citep{RichardsonHMS20}\\
    
    
    \midrule
        COORD & Logical inference about coordinating conjunctions (e.g., & \citep{saha-etal-2020-conjnli}, Coord. \citep{kim-etal-2019-probing}\\
        & \textit{dan}, \textit{tetapi}, \textit{atau}) that conjoin two or more conjuncts of varied &  Coordinations \citep{analyzingANLI}\\
        & syntactic categories (i.e., noun phrases, verb phrases, clauses) & Boolean \citep{joshi-etal-2020-taxinli}\\

    \midrule
        QUANT &	Inferences from the natural language analogs of universal and & Quantification \citep{wang-etal-2018-glue}\\ 
        & existential quantification & Quantifier \citep{joshi-etal-2020-taxinli}\\
    
    
    \midrule    
        COMP & Comparatives and superlatives expressing qualitative or & Comp. \& Super. \citep{analyzingANLI}\\
        & quantitative differences between entities & Comp. \citep{kim-etal-2019-probing}\\
        & & Comparatives \citep{RichardsonHMS20}\\
    
    
    \midrule    
        LEXSEM & Inferences made possible
by lexical information about &    
    Lexical Entailment \citep{wang-etal-2018-glue} \\
        & synonyms, antonyms, and hypernym-hyponyms & Lexical \citep{BentivogliCDGLM10}\\
        & & \citep{glockner-etal-2018-breaking} \\
        & & Lexical \citep{joshi-etal-2020-taxinli}\\
    
    
    \midrule    
        NUM & Numerical expression, such as cardinal and/or ordinal numbers, & \citep{ravichander-etal-2019-equate}\\
        & percentage and money. It also includes the mathematical & Numeral Reasoning \citep{naik-etal-2018-stress}\\
        &  reasoning, and counting of the entities. & Counting \citep{kim-etal-2019-probing}\\
        & & Numeral \citep{analyzingANLI}\\
    

    \midrule
        COREF &	Anaphora and coreferences between pronouns, proper names, & Anaphora/Coreference \citep{wang-etal-2018-glue}\\
        & (e.g. named entities) and noun phrases & Coreference \citep{joshi-etal-2020-taxinli}\\
        & & Reference \citep{analyzingANLI} \\

    
    \midrule
        IDIOM & Idiomatic expression. & Idioms \citep{analyzingANLI}\\
       
        
        \midrule
        SPAT & Spatial reasoning that involves places and spatial relations & Spatial \citep{kim-etal-2019-probing} \\
        & between entities; understanding the preposition of location & Spatial \citep{joshi-etal-2020-taxinli} \\ 
        & and direction & \\

    
        \midrule
        TEMP & Temporal reasoning that involves a common sense of time, & \citep{vashishtha-etal-2020-temporal}\\
        & for example, the duration an event lasts, the general time & \\
        &  an activity is carried out and, the sequence of events\\

    
    \midrule
        CS & Commonsense knowledge that is expected to be possessed by & Common sense \citep{wang-etal-2018-glue}\\
        & most people, independent of cultural or educational background. & Plausibility \citep{analyzingANLI}\\
        & This includes a basic understanding of physical and social & \\
        & dynamics, plausibility of events, and cause-effect relations & \\

    
        \midrule
        WORLD & Reasoning that requires knowledge about named entities, & World knowledge \citep{wang-etal-2018-glue}\\
        & knowledge about historical and cultural, current events; and & Reasoning-Fact \citep{analyzingANLI}\\
        & domain-specific knowledge. & World \citep{joshi-etal-2020-taxinli} \\
        

    \bottomrule
    \end{tabular}
    \caption{The list of linguistic tags in IndoNLI diagnostic set and reference to similar tags in previous work.}
    \label{tab:phenomena}
\end{table*}

\newpage

\section{IndoNLI Diagnostic Set Examples}

\begin{table*}[ht!]
\small
    \centering
    \begin{tabular}{p{0.42\linewidth}  p{0.25\linewidth}  p{0.03\linewidth}  p{0.08\linewidth}}


\toprule
        \textbf{Premise} & \textbf{Hypothesis} & \textbf{Label} & \textbf{Phenomena} \\

    \midrule

Topan   Molave telah menewaskan 36 orang dan menyebabkan 46 orang lainnya hilang di   Vietnam. & Angka kematian lebih sedikit   ketimbang angka orang hilang. & E & COMP,  NUM \\
\textit{(Typhoon Molave has killed 36 people and left 46 others missing in Vietnam.)} & \textit{(The death rate is less than the   number of missing persons.)} &  &  \\

    \midrule

Selain   rumah yang rusak, area persawahan milik warga 5 hektare longsor dengan   kedalaman 10 meter dan lebar 250 meter. & Sawah dan rumah warga mengalami   kerusakan. & E & COORD, MORPH, STRUCT \\
\textit{(In   addition to the damaged houses, a 5-hectare rice field owned by residents had   a landslide with 10 meters depth and 250 meters width.)} & \textit{(Rice fields and houses were   damaged.)} &  &  \\

    \midrule
Penyerang   Juventus, Cristiano Ronaldo, merayakan gol ke gawang Cagliari, Minggu   (22/11/2020). & Ronaldo melakukan perayaan atas   gol yang ia buat. & E & COREF,  MORPH \\
\textit{(Juventus   striker, Cristiano Ronaldo, celebrates a goal against Cagliari, Sunday   (11/22/2020).)} & \textit{(Ronaldo celebrates the goal he   made.)} &  &  \\

    \midrule
Semua   calon petahana Pilkada 2020 di 3 kabupaten di Yogyakarta dinyatakan kalah   dalam rapat pleno penghitungan suara Komisi Pemilihan Umum (KPU). & Ada calon petahana Pilkada 2020   di 3 kabupaten di Yogyakarta yang menang. & C & QUANT, LEXSEM \\
\textit{(All   incumbent candidates for the 2020 Pilkada (regional election) in 3 districts   in Yogyakarta were declared defeated in the plenary meeting of the General   Election Commission (KPU) vote counting.)} & \textit{(There is incumbent candidate for   the 2020 Pilkada in 3 districts in Yogyakarta who won.)} &  &  \\

    \midrule
Kebijakan   untuk membuka sekolah dikembalikan kepada pemerintah daerah dengan   persetujuan orangtua. & Pemerintah daerah dapat membuka   sekolah tanpa persetujuan orang tua. & C & NEG, STRUCT \\
\textit{(The   policy to open schools is assigned to the local government with parental   permission.)} & \textit{(Local governments can open   schools without parental permission.)} &  &  \\

    \midrule
Kota   Gunungsitoli terletak di Pulau Nias dan berjarak sekitar 85 mil laut dari   Kota Sibolga. & Kota Sibolga terletak di Pulau   Nias. & C & SPAT, STRUCT \\
\textit{(Gunungsitoli   City is located on Nias Island and is about 85 nautical miles from Sibolga   City.)} & \textit{(Sibolga City is located on Nias   Island.)} &  &  \\

    \midrule
Perlahan-lahan,   keluarga Kim berusaha agar satu per satu anggota keluarga mereka dapat   bekerja di keluarga Park, dengan saling merekomendasikan satu sama lain dan   berbohong sebagai penyedia jasa profesional yang saling tidak kenal. & Tipu daya keluarga Kim berujung   di meja hijau. & N & IDIOM, CS \\
\textit{(Gradually,   the Kims try to get each of their family members to work for the Parks,   recommending each other and lying as professional service providers who don't   know each other.)} & \textit{(The Kim family's trickery ended   up at the court.)} &  &  \\

    \midrule
Ismed   Sofyan (lahir di Manyak Payed, Aceh Tamiang, 28 Agustus 1979; umur 41 tahun)   adalah pemain Persija dan tim nasional Indonesia. & Ismed Sofyan menghabiskan masa   mudanya di Jakarta. & N & TEMP, WORLD \\
\textit{(Ismed   Sofyan (born in Manyak Payed, Aceh Tamiang, August 28, 1979; age 41) is a   Persija player and the Indonesian national team.)} & \textit{(Ismed Sofyan spent his youth in   Jakarta.)} &  & \\
\bottomrule
\end{tabular}

\caption{Annotated examples from the diagnostic set}
\label{tab:last-table}
\end{table*}

\twocolumn

\end{document}